# Human-AI Co-Creation Approach to Find Forever Chemicals Replacements

JULIANA JANSEN FERREIRA, IBM Research, Brazil
VINÍCIUS SEGURA, IBM Research, Brazil
JOANA G. R. SOUZA, IBM Research, Brazil
GABRIEL D. J. BARBOSA, IBM Research, Brazil
JOÃO GALLAS, IBM Research, Brazil
RENATO CERQUEIRA, IBM Research, Brazil
DMITRY ZUBAREV, IBM Research, USA

Generative models are a powerful tool in AI for material discovery. We are designing a software framework that supports a human-AI co-creation process to accelerate finding replacements for the "forever chemicals"– chemicals that enable our modern lives, but are harmful to the environment and the human health. Our approach combines AI capabilities with the domain-specific tacit knowledge of subject matter experts to accelerate the material discovery. Our co-creation process starts with the interaction between the subject matter experts and a generative model that can generate new molecule designs. In this position paper, we discuss our hypothesis that these subject matter experts can benefit from a more iterative interaction with the generative model, asking for smaller samples and "guiding" the exploration of the discovery space with their knowledge.

CCS Concepts: • **Human-centered computing** → **Human computer interaction (HCI)**.

Additional Key Words and Phrases: generative models, human-in-the-loop, domain expert

## 1 INTRODUCTION

Generative models (GMs) are the current state-of-the-art in Artificial Intelligence (AI) techniques for material discovery due to their ability to generate large volumes of novel molecules across different domains. These models exhibit extreme "creativity" that often translates to low viability of the generated candidates, requiring a human-in-the-loop stage to filter the results. This human-in-the-loop stage has two intertwined challenges:

(1) a *data exploration challenge*, since the interaction with all molecules generated by the GM (solution space) would be too overwhelming for users and a significant user experience challenge [8, 11], and
(2) an *user interaction challenge*, since the strategy of using smaller samples would be a way to guide users in the exploration of the discovery space (all possible – known and unknown – solutions) by looking at it in parts, building their solution space in each iteration of GM [2, 7, 11].

We are designing Discovery Workbench (DWb), a software framework that supports a process of human-AI co-creation to accelerate finding replacements for PFAS materials, known as "forever chemicals" because they do not break down over time and thus can remain permanently in the air, soil, water and in the human body [5]. We envision empowering subject matter experts (SMEs) to work with the GM's output, combining it with their domain-specific tacit knowledge to focus the discovery process on *feasible* and potential candidates *for a given application* [5].

Our system is not focused on providing users with some level of understanding about the inner functioning of generative models, like many previous works [1, 3, 11]. Instead, we focus on ways in which SMEs can interact with the GM themselves and influence its functioning in significant ways.

In our co-creation process, the SMEs can influence how the GM will generate the candidate molecules by informing different parameters (like target molecule) and interpreting the results with their domain-specific tacit knowledge.

In this position paper, we discuss our hypothesis that the SME can identify potential candidates earlier, using less computational resources, and with more trust in the GM if they have more iterations with it (*i.e.*, running it several times), producing smaller samples. We are planning to test this hypothesis in user studies with SMEs from the PFAS domain.

## 2  OUR PFAS REPLACEMENT DISCOVERY CASE

The replacement of PFAS substances, also know as "forever chemicals," is a high-profile application of generative systems for materials discovery. Since they are bioaccumulative and toxic, with high levels of use resulting in detectable levels found in the bloodstream of most humans [17], regulatory agencies are working to approve new restrictions for PFAS's use. This makes their replacement not only a challenge of environmental and health protection, but a requirement for regulatory compliance [4, 18].

The centrality of certain PFAS uses in modern life makes replacement particularly challenging, with over 200 application spaces in need of viable replacements, many consisting of very large scale applications. In many cases, the material ultimately ends up in the ocean, being widely present in rainwater and waterways worldwide as a result [6].

Taking into account the current state of technological advancement, we are ill-equipped to meet the demand for PFAS replacements in the above-cited levels. In recent years, GMs appeared as one of the promising ways to help SMEs make the material discovery process faster and easier [10, 12, 14, 15, 20]. These models can generate new molecule designs based on properties such as target proteins, target omics profiles, scaffolds distances, binding energies, and additional targets relevant for materials and drug discovery [13].

## 3  OUR HUMAN-AI CO-CREATION SCENARIO

In order to accelerate the discovery of new alternatives to PFAS materials, SMEs may take advantage of generative models to be able to generate potentially promising candidates. We are designing Discovery Workbench (DWb), a software framework that supports a process of human-AI co-creation to accelerate finding replacements for PFAS materials [5]. In our work, SMEs co-create molecular designs for PFAS replacement alongside GT4SD[1], an open-source library to accelerate hypothesis generation in the scientific discovery process. These models, however, are complex and using them in a way that takes advantage of the SMEs' domain-specific tacit knowledge is not trivial.

It is important to note that our use case does not involve users that need to understand the inner logic of the GM involved (such as an AI engineer or data scientist). These SMEs only need to be able to interact with the GM and, most importantly, interpret the results during the discovery process. This combination is a key difference from other common solutions: we are focusing on people with domain knowledge that interact with AI in their contextualized task, as opposed to just training an AI tool. They possess significant domain knowledge and see the GMs as instruments in their exploration of the latent space of interest and not as their main object of interest.

In figure 1, we present an example of what we call *the SME iterative interaction with GM*. Instead of executing the GM only once and generating a vast amount of samples that have to be filtered by other means afterwards, in our design users interact with the model iteratively, generating smaller samples each time. As we can see in figure 1, in the 1st iteration, the SME provides the GM a target molecule (which they may have obtained from literature, for example) to be used as a reference point for the generation of candidate molecules, as well as other parameters (such as the number of samples they desire). The SME would then inspect the results of this 1st iteration, bringing their domain-knowledge to the process, and select one of the generated molecules as the target molecule for a second iteration. This process could be repeated as many times as the expert desires. However, they may eventually select another target molecule, not generated in any previous iteration, thereby initiating a new graph of potential candidates (exemplified by the 4th iteration in the figure). Through this process of extending previous targets or creating new ones, the SME can explore the discovery space, taking advantage of their domain knowledge to prioritize which avenues to explore further.

---

[1]Generative Toolkit 4 Scientific Discovery: https://github.com/GT4SD

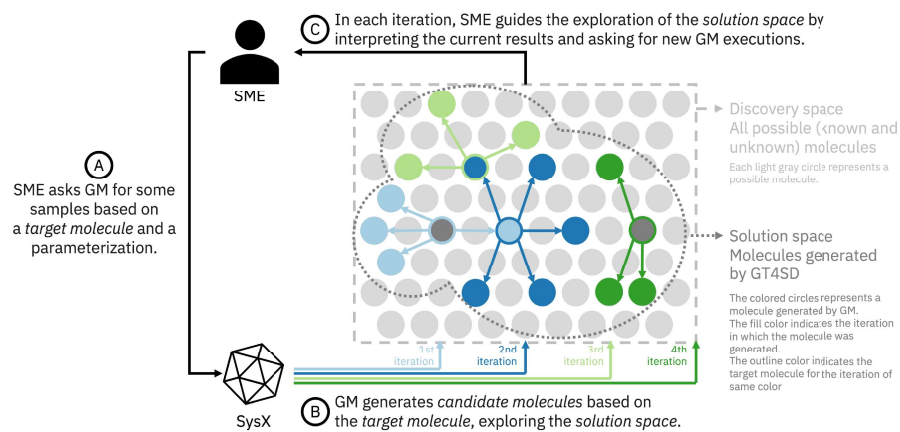

Fig. 1. SME iterative interactions with GT4SD

The key idea with this iterative interaction approach is to allow SMEs to guide the exploration of the discovery space, creating a more tailored solution space. To tackle the aforementioned *data exploration challenge*, we can augment the GM results in DWb with knowledge that is specific to the SMEs' problem, thus improving their capacity to interpret the results by adding extra visualization and signification layers. Instead of having them face an enormous sample of candidates which may or may not fit their needs, we allow them to deal with them in batches, selecting the most promising to serve as the basis for the next batch of generated candidates and potentially ignoring the rest, thus targeting the *user interaction challenge*.

Moreover, having a tighter expert-in-the-loop approach allows SMEs to interact with intermediate results (considering the whole process). On the one hand, from the DWb perspective, it means more decision-making points – key events in which knowledge can be captured to be later reused by the system (*e.g.*, in training new models). On the other hand, from the SMEs perspective, it means waiting less time for results that are better suited to theirs goals. We also believe that, by investing into the human-AI co-creation approach, it may lead to an increase in the perceived GM trustworthiness by the SMEs, even without providing any additional explainability to its "black box". Our hypothesis (to be considered in future studies) is that, by having results that are more easily understandable (by reducing its size and providing new ways to explore) and by providing more "checkpoints" in the overall process (thus increasing the SME feedback of the process), SMEs will have greater trust in the results and ownership of the process – moving from a simple expert-in-the-loop scheme to true human-AI co-creation.

It is important to note that these advantages rely on the SMEs understanding of the domain in order to improve the exploratory process. Individuals lacking in this respect would not be able to differentiate between the generated candidates and guide the GM effectively. In complex domains, such as that of molecular design for PFAS replacement, domain knowledge is essential.

In our case, this exploration of potential alternatives for PFAS materials serves only as a starting point. Once this dataset is formed, these candidates are investigated further in a variety of ways (*e.g.*, other triage features, simulations, risk assessment). Therefore, having a set of candidates tailored by the SME that is going to be conducting these investigations is ideal. By having some control over the exploration of the discovery space, we also posit that they might have the opportunity to identify potential candidates much earlier in the process.

## 4 FINAL REMARKS

Our goal with this position paper was to present our proposal for a human-AI co-creation process, considering iterative interactions between SMEs and GMs. In this co-creation process, SMEs usually have little understanding of the behavior of the GM, only having enough to decide which way to go during that process (*i.e.*, parameterize GM executions). There are still several challenges in the literature in explaining GM behavior to users that have little knowledge about algorithms and GMs. Some examples are the challenges related to interpretability [9, 12], global view of the model [16], disentanglement and reconstruction quality of results [3], high-dimensional datasets, large number of iterations before convergence, and challenges for cooperative use by experts [11].

In our case, where "looking at the insides" of the GM is not an option for our user (the SME), our hypothesis considers only two elements: *the input*, what the SME can inform the GM; and *the output*, what the GM generates (sample molecules) from that input. By having a smaller human-in-the-loop cycles, SMEs has more chances to give feedback to the GM, thus imprint their domain-specific tacit knowledge in the process – for example, by the selection of the target molecule in the *input*, which can be a proxy for their discovery goals when interacting with GM. Additionally, the *output* may be investigated by the SMEs using domain-specific tools. Using this approach, we believe that SMEs can identify potential candidates for PFAS replacement:

- earlier in the process – since there are smaller human-in-the-loop cycles, the final outcome of the GM step is more tailored to the SMEs' discovery goals;
- using less computation resources – since each iteration generates less samples and in a more "targeted" exploration, the overall "cost" of running the GM would be reduced; and
- with more trust in the GM – since the SME and the GM would be working closely together, this would increase the SME sense of ownership of the overall process [19].

We believe this SME-GM partnership can offer many opportunities for material discovery. As future research, we have a central research question: *Can SME users learn the model's behavior through iterative interactions with a specific goal in mind?* If so: *How can we deal with the challenges involved in these iterative interactions with GMs when they can "change their behavior" across different interactions?*. These questions will guide our next steps in designing user studies with SMEs regarding the PFAS replacement problem. Even if our hypothesis is not confirmed, we would better understand how SMEs perceive and interact with GMs, serving to inform our future research and development of Human-AI co-creation processes.